\documentclass[11pt]{article}
\usepackage{arabtex}
\usepackage{coling2018}
\usepackage{utf8}
\setcode{utf8}
\usepackage{url}
\usepackage{graphicx} % use \rotatebox{90}{rota}
\usepackage[stable]{footmisc}
\date{}

\begin{document}

\title{\textbf{Can Linguistic Distance help Language Classification?\\Assessing Hawrami-Zaza and Kurmanji-Sorani}}

\author{
	\begin{tabular}[t]{c}
		Hossein Hassani\\
		\textnormal{University of Kurdistan Hewl\^er}\\
		\textnormal{Kurdistan Region - Iraq}\\
		{\tt hosseinh@ukh.edu.krd}
	\end{tabular}
}

\maketitle

\begin{abstract}
To consider Hawrami and Zaza (Zazaki) standalone languages or dialects of a language have been discussed and debated for a while among linguists active in studying Iranian languages. The question of whether those languages/dialects belong to the Kurdish language or if they are independent descendants of Iranian languages was answered by \newcite{mackenzie1961origins}. However, a majority of people who speak the dialects are against that answer. Their disapproval mainly seems to be based on the sociological, cultural, and historical relationship among the speakers of the dialects. While the case of Hawrami and Zaza has remained unexplored and under-examined, an almost unanimous agreement exists about the classification of Kurmanji and Sorani as Kurdish dialects. The related studies to address the mentioned cases are primarily qualitative. However, computational linguistics could approach the question from a quantitative perspective. 
In this research, we look into three questions from a linguistic distance point of view. First, how similar or dissimilar Hawrami and Zaza are, considering no common geographical coexistence between the two. Second, what about Kurmanji and Sorani that have geographical overlap. Finally, what is the distance among all these dialects, pair by pair? We base our computation on phonetic presentations of these dialects (languages), and we calculate various linguistic distances among the pairs. We analyze the data and discuss the results to conclude\footnote{This paper is based on an abstract that was submitted to and presented under the same title at the {\textit{5\textsuperscript{th} International Conference on Kurdish Linguistics}} at the University of Graz, 25 September 2021.}.
\end{abstract}

\section{Introduction}
\label{sec:intro}

Kurdish is an Indo-European language that specifically belongs to the Iranian languages, and more particularly to Western-Iranian \cite{kreyenbroek2005kurdish}, or North-Western-Iranian languages \cite{minorsky1945tribes,phillipson1996colonial}. The language is multi-dialect \cite{khalid2015kurdish}, but there is no consensus about the dialects among the scholars. While according to \newcite{mackenzie1961origins}, Hawrami and Zaza (Zazaki) are standalone languages that he considers directly under Western-Iranian languages, \cite{hassanpour1992nationalism} addresses them as dialects of Kurdish. The question of which one of the \textit{commonly} known dialects are should be recognized as  Kurdish from a linguistic perspective has been an ongoing debate among the scholars who have studied the language for a long while. Importantly, a majority of people who speak the dialects believe that all of those dialects/languages are Kurdish. Their belief mainly seems to be based on the sociological, cultural, and historical relationship among the speakers of the dialects. 

While the case of Hawrami and Zaza has remained unexplored and under-examined, an almost unanimous agreement exists about the classification of Kurmanji and Sorani as Kurdish dialects. The related studies to address the mentioned cases are primarily qualitative. However, computational linguistics could approach the question from a quantitative perspective.
 
In this research, we look into three questions from a linguistic distance point of view. First, how similar or dissimilar Hawrami and Zaza are, considering no common geographical coexistence between the two. Second, what about Kurmanji and Sorani that have geographical overlap. Finally, what is the distance among all these dialects, pair by pair? We base our computation on phonetic presentations of these dialects (languages), and we calculate various linguistic distances among the pairs.

The rest of this paper is organized as follows. Section \ref{sec:rw} reviews the related work. In Section \ref{sec:method}, we present the method that we follow to compute the linguistic distance. We report and discuss the results in Section~\ref{sec:result}. Finally, Section \ref{sec:conc} concludes the paper and suggests some areas for future work.

\section{Related Work}
\label{sec:rw}

\newcite{mackenzie1961origins} set the basis of Kurdish language classification. Their work is still one of the essential references for Kurdish studies. Their qualitative approach was combined with historical studies of Kurdish development by other scholars such as \newcite{hassanpour1992nationalism}. While \newcite{mackenzie1961origins} classified Kurmanji and Sorani under Kurdish stem, he considered Hawrami and Zaza as descendants of Iranian languages but not Kurdish.

\newcite{minorsky1943guran} discussed Gurani from a historical point of view. Although he presented and discussed some literature written in Gurani, his view was primarily based on historical analysis. \newcite{minorsky1943guran} considered Zaza and “Awrami” (Hawrami) as dialects of Gurani, which was claimed again by some other scholars. All those claims are mostly based on qualitative measures as well. However, several scholars argued against their proposition (see \newcite{hassanpour1993kurdish}), 1993, for example). \newcite{mackenzie1991compendium} provided one of the earliest critiques on the matter. The issue was recently addressed again by \newcite{haig2014introduction} and \newcite{haig2018languages}. But, the classification of those dialects (languages) is still an open question. Regardless of their proposed classifications, so far, Kurdish linguists have followed a qualitative approach, and no quantitative study has been conducted about the subject.

However, research on language classification according to quantitative methods, for example, linguistic distance, has been emerging. To illustrate, \newcite{gamallo2017language} classified European languages based on their similarities and divergence according to their linguistic distance. Also, \newcite{rama2015comparative} suggested a genetic classification of languages based on Swadesh-style core vocabulary. Using Swadesh's list \cite{swadesh1955towards} as a basis for language similarity has been practiced for a long while (see, for instance, \newcite{holman2008advances}).
In this study, we also use the Swadesh list as the fundamental instrument.

\section{Method}
\label{sec:method}

Our method is similar to what is proposed by \cite{bourgeoislingua}. We prepare a Swadesh list (a 207 entry version) for Zaza, Hawrami, Kurmanji, and Sorani. We calculate the Jaro similarity/distance between Zaza and Hawrami on the one hand and Kurmanji and Sorani on the other. We also calculate the linguistic distance between all mentioned dialects. We then discuss the finding against the previous qualitative work to investigate the conformance to or divergence of their propositions about the classification of the mentioned dialects (languages).

\section{Result and Discussion}
\label{sec:result}

We prepare a Swadesh list (a 207 entry version) for each dialect under study\footnote{The dataset is available at {\url{https://github.com/KurdishBLARK/Dialect-Classification}}}. Figure \ref{tab:sim-dist} shows the similarities and distances between each pair of the dialects. Figure \ref{tab:wsim-dist} shows the number of words that were completely similar or were completely different between the Kurdish dialects. Figure \ref{tab:wsim-dist} shows the number of words that were completely similar or were completely different between the Kurdish dialects.

\begin{table}[ht]
	\centering
	\caption{Similarity/Distance between Kurdish Dialects (Za: Zaza; Hwa: Hawrami; Kur: Kurmanji; Sor: Sorani).}
    \label{tab:sim-dist}
	\scalebox{1}{
		\begin{tabular}{|c|c|c|c|c|c|c|c|}
			\hline
			{Jaro (Avg)} & Za-Haw & Kur-Sor & Za-Kur& Za-Sor& Haw-Kur& Haw-Sor \\ \hline \hline 
			Similarity& 0.57 & 0.68 & 0.59 & 0.58 & 0.52 & 0.61 \\ \hline
			Distance  & 0.43 & 0.32 & 0.41 & 0.42 & 0.48 & 0.39 \\ \hline
		\end{tabular}
	}
\end{table}

\begin{table}[ht]
	\centering
	\caption{Number of Completely Similar or Different between Kurdish Dialects (Za: Zaza; Hwa: Hawrami; Kur: Kurmanji; Sor: Sorani).}
	\label{tab:wsim-dist}
	\scalebox{1}{
		\begin{tabular}{|c|c|c|c|c|c|c|c|}
			\hline
			Similarity & Za-Haw & Kur-Sor & Za-Kur& Za-Sor& Haw-Kur& Haw-Sor \\ \hline \hline 
			Completely Similar   & 10 & 55 & 22 & 19 &  7 & 23  \\ \hline
			Completely Different & 20 & 23 & 29 & 32 & 37 & 21  \\ \hline
		\end{tabular}
	}
\end{table}

\begin{table}[ht!]
	\centering
	\caption{Percentages of Completely Similar or Different Words between Kurdish Dialects (Za: Zaza; Hwa: Hawrami; Kur: Kurmanji; Sor: Sorani).}
	\label{tab:wpsim-dist}
	\scalebox{1}{
		\begin{tabular}{|c|c|c|c|c|c|c|c|}
			\hline
			Similarity & Za-Haw & Kur-Sor & Za-Kur& Za-Sor& Haw-Kur& Haw-Sor \\ \hline \hline 
			Completely Similar    & 4.83\% & 26.57\% & 10.63\% &  9.18\% &  3.38\% & 11.11\% \\ \hline
			Completely Different  & 9.66\% & 11.11\% & 14.01\% & 15.48\% & 17.87\% & 10.14\% \\ \hline
		\end{tabular}
	}
\end{table}

We observe that Zazaki and Hawrami are less similar than Kurmanji and Sorani, while
Zazaki and Hawrami share less number of completely similar words than Kurmanji and Sorani. Also, Hawrami and Sorani are less similar than Kurmanji and Sorani, but the difference is not significant. Zazaki and Sorani are as different as Zazaki and Hawrami, while Hawrami and Kurmanji are as similar as they are different, though the similarity weighs more. Finally, the similarity of Zazaki – Kurmanji and Zazaki – Sorani is almost the same.

The results do not suggest that Zazaki and Hawrami are any closer than Kurmanji and Sorani. Therefore, some suggestion that considers them as a single language/dialect cannot be attested. Furthermore, the results show that Hawrami and Kurmanji are quite similar to Kurmanji and Sorani that could open a discussion among the scholars who consider these dialects not to be Kurdish to reconsider the way that they have classified those dialects.

\section{Conclusion}
\label{sec:conc}
In this research, we looked into the similarity and distances of Kurdish dialects. We prepared Swadesh lists (the 207 entry version) for each one of the four Kurdish dialects, namely Kurmanji, Sorani, Hawrami, and Zazaki. To calculate the similarity and difference between each pair of those dialects, we applied the Jaro measure. The results did not suggest that Zazaki and Hawrami are any closer than Kurmanji and Sorani. That suggests a need to investigate and perhaps reconsider the way that those dialects have been classified.

In the future, we would like to expand the dataset beyond the Swadesh list to a larger common vocabulary that could help to obtain a more thorough view of the similarity/distance between Kurdish dialects. Also, we are interested in considering other lists such as Leipzig–Jakarta and comparing the results with the current ones to see whether they are reciprocal or not.

\section*{Acknowledgments}

My utmost appreciations to \c{S}ad\^e Agon for her generous assistance in preparing Zazaki Swadesh list, Dr. Srwa Rostaminezhad for her generous assistance in preparing Hawrami Swadesh list, S\^ipan Monnet, Saman Idrees, and Marwan Ali for their generous assistance in preparing Kurmanji Swadesh list, and Dr. Ergin {\"O}pengin and Dr. Nava Hassani for their help in coordination with the Zazaki and Hawrami informants 

\bibliographystyle{lrec}
\bibliography{KurdishLnagClass}

\begin{thebibliography}{}

\bibitem[\protect\citename{Bourgeois-Gironde \bgroup et al.\egroup
  }2021]{bourgeoislingua}
Bourgeois-Gironde, S., Ginsburgh, V., Hassani, H., and Weber, S.
\newblock (2021).
\newblock {A Lingua Franca for Kurdish Populations}.
\newblock {London, Centre for Economic Policy Research}
  \url{https://cepr.org/active/publications/discussion_papers/dp.php?dpno=16086}.

\bibitem[\protect\citename{Gamallo \bgroup et al.\egroup
  }2017]{gamallo2017language}
Gamallo, P., Pichel, J.~R., and Alegria, I.
\newblock (2017).
\newblock From language identification to language distance.
\newblock {\em Physica A: Statistical Mechanics and its Applications},
  484:152--162.

\bibitem[\protect\citename{Haig and Khan}2018]{haig2018languages}
Haig, G. and Khan, G.
\newblock (2018).
\newblock {\em The Languages and Linguistics of Western Asia: An Areal
  Perspective}, volume~6.
\newblock Walter de Gruyter GmbH \& Co KG.

\bibitem[\protect\citename{Haig and {\"O}pengin}2014]{haig2014introduction}
Haig, G. and {\"O}pengin, E.
\newblock (2014).
\newblock Introduction to special issue-kurdish: A critical research overview.
\newblock {\em Kurdish studies}, 2(2):99--122.

\bibitem[\protect\citename{Hassanpour}1992]{hassanpour1992nationalism}
Hassanpour, A.
\newblock (1992).
\newblock {\em Nationalism and language in Kurdistan, 1918-1985}.
\newblock San Francisco: Mellen Research University Press.

\bibitem[\protect\citename{Hassanpour}1993]{hassanpour1993kurdish}
Hassanpour, A.
\newblock (1993).
\newblock Kurdish studies: Orientalist, positivist, and critical approaches:
  Review essay.

\bibitem[\protect\citename{Holman \bgroup et al.\egroup
  }2008]{holman2008advances}
Holman, E.~W., Wichmann, S., Brown, C.~H., Velupillai, V., M{\"u}ller, A.,
  Bakker, D., et~al.
\newblock (2008).
\newblock Advances in automated language classification.
\newblock {\em Quantitative Investigations In Theoretical Linguistics (QITL3)},
  page~40.

\bibitem[\protect\citename{Khalid}2015]{khalid2015kurdish}
Khalid, H.~S.
\newblock (2015).
\newblock Kurdish dialect continuum, as a standardization solution.
\newblock {\em International Journal of Kurdish Studies}, 1(1):27--39.

\bibitem[\protect\citename{Kreyenbroek}2005]{kreyenbroek2005kurdish}
Kreyenbroek, P.~G.
\newblock (2005).
\newblock On the kurdish language.
\newblock In {\em The Kurds}, pages 62--73. Routledge.

\bibitem[\protect\citename{MacKenzie}1961]{mackenzie1961origins}
MacKenzie, D.~N.
\newblock (1961).
\newblock {The Origins Of Kurdish}.
\newblock {\em Transactions of the Philological Society}, 60(1):68--86.

\bibitem[\protect\citename{MacKenzie}1991]{mackenzie1991compendium}
MacKenzie, D.
\newblock (1991).
\newblock {Compendium Linguarum Iranicarum}.

\bibitem[\protect\citename{Minorsky}1943]{minorsky1943guran}
Minorsky, V.
\newblock (1943).
\newblock {The Guran}.
\newblock {\em Bulletin of the School of Oriental and African Studies}, 11(Part
  I):75--103.

\bibitem[\protect\citename{Minorsky}1945]{minorsky1945tribes}
Minorsky, V.
\newblock (1945).
\newblock {The tribes of western Iran}.
\newblock {\em The Journal of the Royal Anthropological Institute of Great
  Britain and Ireland}, 75(1/2):73--80.

\bibitem[\protect\citename{Phillipson and
  Skutnabb-Kangas}1996]{phillipson1996colonial}
Phillipson, R. and Skutnabb-Kangas, T.
\newblock (1996).
\newblock {Colonial language legacies: the prospects for Kurdish}.
\newblock In {\em Self-determination: international perspectives}, pages
  200--213. Palgrave Macmillan.

\bibitem[\protect\citename{Rama and Borin}2015]{rama2015comparative}
Rama, T. and Borin, L.
\newblock (2015).
\newblock Comparative evaluation of string similarity measures for automatic
  language classification.
\newblock In {\em Sequences in language and text}, pages 171--200. De Gruyter
  Mouton.

\bibitem[\protect\citename{Swadesh}1955]{swadesh1955towards}
Swadesh, M.
\newblock (1955).
\newblock Towards greater accuracy in lexicostatistic dating.
\newblock {\em International journal of American linguistics}, 21(2):121--137.

\end{thebibliography}

\end{document}